\documentclass[conference]{IEEEtran}
\IEEEoverridecommandlockouts
\usepackage{cite}
\usepackage{amsmath,amssymb,amsfonts}
\usepackage{graphicx}
\usepackage{textcomp}
\usepackage{xcolor}
\usepackage{amsthm}
\usepackage{xcolor}
\usepackage{graphicx}
\usepackage{multirow}
\usepackage{array}
\usepackage{xspace}
\usepackage{enumitem}
\usepackage{mathtools}
\usepackage[ruled,linesnumbered]{algorithm2e}
\usepackage{algpseudocode} 
\setlist[itemize]{leftmargin=*}
\setlist[enumerate]{leftmargin=*}
\usepackage{bm}
\usepackage{hyperref}
\usepackage{booktabs}
\usepackage{makecell}

\newtheorem{problem}{Problem}

\def\BibTeX{{\rm B\kern-.05em{\sc i\kern-.025em b}\kern-.08em
    T\kern-.1667em\lower.7ex\hbox{E}\kern-.125emX}}
   
\begin{document}

\title{Uncertainty-aware Traffic Prediction \\under Missing Data
}

\author{\IEEEauthorblockN{Hao Mei}
\IEEEauthorblockA{\textit{School of Computing and Augmented Intelligence} \\
\textit{Arizona State University}\\
Tempe, US \\
hmei7@asu.edu} \\

\IEEEauthorblockN{Zhiming Liang}
\IEEEauthorblockA{\textit{Faculty of Electronic and Information Engineering} \\
\textit{Xi'an Jiaotong University}\\
Xi'an, China\\
zhimingliang@stu.xjtu.edu.cn} \\

\IEEEauthorblockN{Bin Shi}
\IEEEauthorblockA{\textit{Faculty of Electronic and Information Engineering} \\
\textit{Xi'an Jiaotong University}\\
Xi'an, China\\
shibin@xjtu.edu.cn}

\and

\IEEEauthorblockN{Junxian Li}
\IEEEauthorblockA{\textit{Faculty of Electronic and Information Engineering} \\
\textit{Xi'an Jiaotong University}\\
Xi'an, China \\
ljx201806@stu.xjtu.edu.cn} \\

\IEEEauthorblockN{Guanjie Zheng}
\IEEEauthorblockA{\textit{School of Electronic Information and Electrical Engineering} \\
\textit{Shanghai Jiaotong University}\\
Shanghai, China \\
gjzheng@sjtu.edu.cn} \\

\IEEEauthorblockN{Hua Wei$^{\ast}$\thanks{*Corresponding author.}}
\IEEEauthorblockA{\textit{School of Computing and Augmented Intelligence} \\
\textit{Arizona State University}\\
Tempe, US\\
hwei27@asu.edu}
}

\maketitle

\begin{abstract}
Traffic prediction is a crucial topic because of its broad scope of applications in the transportation domain. Recently, various studies have achieved promising results. However, most studies assume the prediction locations have complete or at least partial historical records and cannot be extended to non-historical recorded locations. In real-life scenarios, the deployment of sensors could be limited due to budget limitations and installation availability, which makes most current models not applicable. Though few pieces of literature tried to impute traffic states at the missing locations, these methods need the data simultaneously observed at the locations with sensors, making them not applicable to prediction tasks. Another drawback is the lack of measurement of uncertainty in prediction, making prior works unsuitable for risk-sensitive tasks or involving decision-making. To fill the gap, inspired by the previous inductive graph neural network, this work proposed an uncertainty-aware framework with the ability to 1) extend prediction to missing locations with no historical records and significantly extend spatial coverage of prediction locations while reducing deployment of sensors and 2) generate probabilistic prediction with uncertainty quantification to help the management of risk and decision making in the down-stream tasks. Through extensive experiments on real-life datasets, the result shows our method achieved promising results on prediction tasks, and the uncertainty quantification gives consistent results which highly correlated with the locations with and without historical data. We also show that our model could help support sensor deployment tasks in the transportation field to achieve higher accuracy with a limited sensor deployment budget. 
\end{abstract}


\section{Introduction}

Traffic prediction is an important problem in urban computing with numerous applications ranging from urban mobility systems to autonomous vehicle operations \cite{zheng2014urban, lana2018road}. In recent years, the availability of large-scale traffic datasets, coupled with advancements in information collection infrastructures, has led to increased attention and analysis of these datasets \cite{cressie2015statistics, jagadish2014big, hoang2016fccf}. Notably, significant progress has been made in traffic prediction accuracy, primarily driven by data-driven approaches and the proliferation of deep learning models. For instance, Ma et al. \cite{ma2017learning} leverage Convolutional Neural Networks to model spatial correlations, while Liu et al. \cite{li2017diffusion} employ Graph Neural Networks to capture spatiotemporal dependencies using the diffusion mechanism.

Despite the promising results achieved in recent studies on traffic prediction, there are two main limitations in most existing works. Firstly, many assume that all prediction locations are equipped with sensors and can access historical data. However, in real-life scenarios, due to budget constraints or limited accessibility, it is common for specific locations of interest to be absent from sensor coverage. Consequently, existing prediction models only apply to locations with historical data observable, limiting prediction capabilities at locations missing historical observation. There is a critical need for models that can forecast observable and missing locations. While some methods have focused on the kriging problem, which involves data imputation using surrounding location data \cite{cressie2015statistics, bahadori2014fast, wu2021inductive}, these approaches require simultaneous access to current data and cannot be directly applied to predict future data.

Secondly, most current traffic prediction studies primarily report the accuracy of their models, neglecting the crucial aspect of prediction uncertainty. In many downstream tasks, especially those involving risk management or decision-making based on prediction results, the quantification of uncertainty is essential in determining whether a method can be implemented in real-life. Failure to assess potential risks can lead to significant costs in high-stakes transportation tasks. Some recent works have attempted to address this issue. For example, Wu et al. \cite{wu2021quantifying} quantified uncertainty in spatiotemporal prediction tasks by discussing different uncertainty quantification methods from frequentist and Bayesian perspectives. Similarly, Prob-GNNs \cite{wang2023uncertainty} proposed a probabilistic graph neural network framework and investigated various probabilistic assumptions specific to ride-sharing demand tasks. However, these methods assume the availability of historical data at each location for evaluating prediction uncertainty. Importantly, locations missing historical observation exhibit distinct topology properties within their surrounding sub-graphs, which locations with historical data may not adequately represent. As a result, predictions at these locations are characterized by low confidence and need to be discerned among all points of interest. Thus, there is an urgent need for a new framework that enables predictions while uncovering the spatiotemporal pattern of uncertainty and extends to locations missing historical observation.

Motivated by the concept of inductive graph neural networks \cite{wu2021inductive}, initially employed to solve spatial kriging problems, we adapt this idea to the context of traffic prediction and propose an \textbf{U}ncertainty-awarded \textbf{I}nductive \textbf{G}raph \textbf{N}eural \textbf{N}etwork (UIGNN) to attack this problem. Our framework can extend to new locations missing historical data while integrating uncertainty quantification methods to assess predictions comprehensively. The details description is shown in Fig.~\ref{fig: framework}. The main contributions of our proposed framework are as follows:

\begin{figure*}[!t]
\centering
  \begin{tabular}{cc}
  \includegraphics[width=0.45\linewidth]{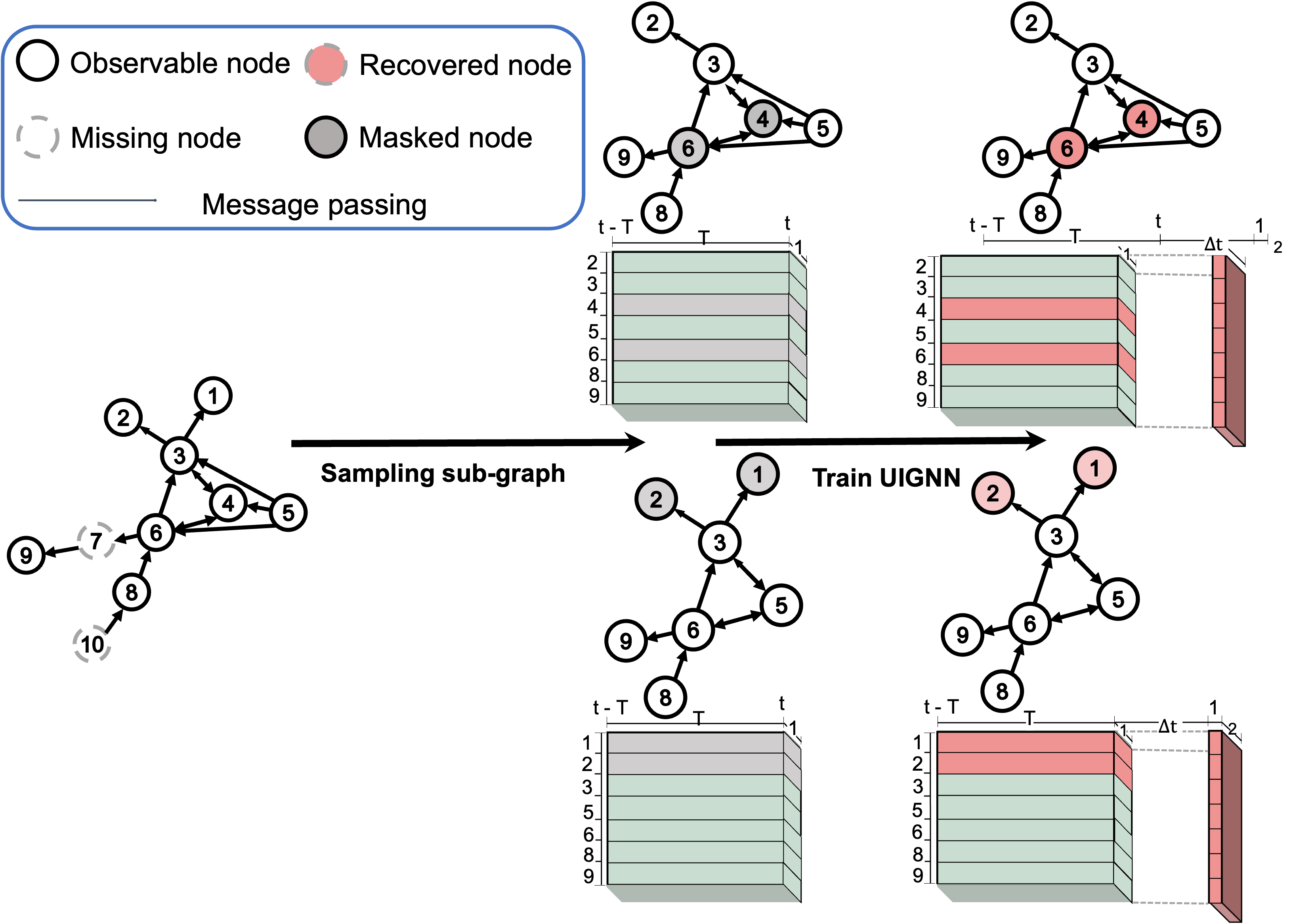} &
   \includegraphics[width=0.45\linewidth]{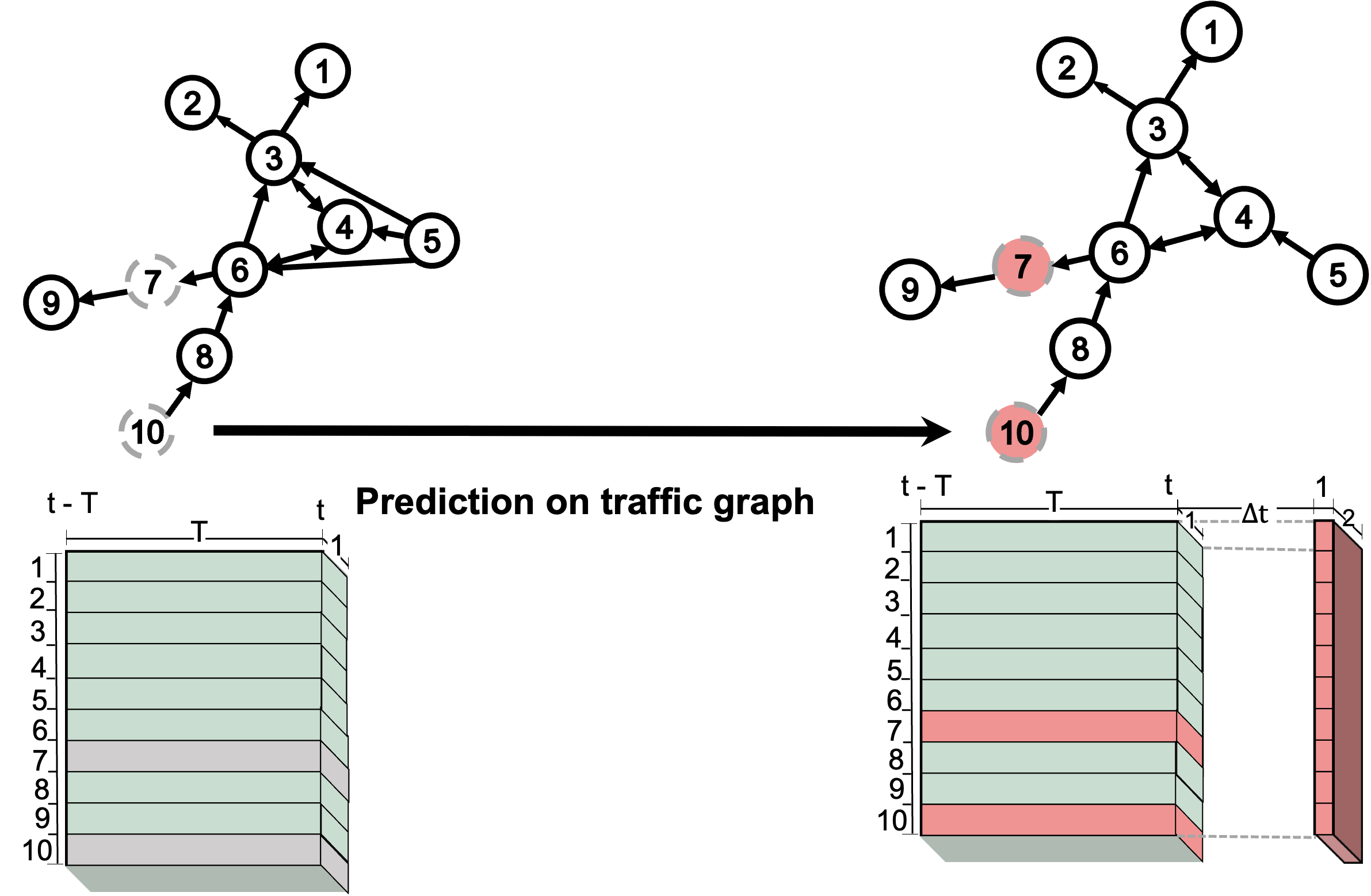}
  \\
  (a) Training procedure of UIGNN
  &

  (b) Prediction procedure of UIGNN
  \end{tabular}
 \caption{Framework of UIGNN. (a) The missing locations indexes are 7,10. And during training, we random sample sub-graph and mask locations 4,6 (upper sample) and train GNN to impute mask locations from time $t-T$ to $t$ and predict all sampled locations at $t+\Delta t$ and quantify the prediction uncertainty. (b) During prediction, UIGNN predicts both missing and observable locations at time $t+\Delta t$ and quantifies the prediction uncertainty.}
 \label{fig: framework}
\end{figure*}

\noindent$\bullet$ Our work is the first to investigate traffic prediction problems concerning the ability to be spatially extended to locations missing historical data in the road network. Our proposed single-step model (directly predict) in traffic prediction problems outperforms the other two-step (impute then predict) methods on the missing data locations. At the same time, keep the same level of performance at observable data locations.

\noindent$\bullet$ To investigate the model performance at different locations, we integrate uncertainty quantification to evaluate the prediction quality at each road network location since the error-based metric is not accessible at missing data locations. The result shows the uncertainty reported by our model can successfully distinguish observable and missing locations and reflect the prediction accuracy. And we also justify our model in the downstream active sensing tasks by using uncertainty to assist deployment of sensors. 

\noindent$\bullet$ We investigate our model on three real-world datasets with historical data observable at  partial locations. The result shows our model performs better than two-step approaches and is effective in real-world sensor deployment tasks.

\section{Related Work}

\subsection{Traffic Prediction}
Traffic prediction has been a challenging problem due to the complex dependencies in the time and space domain. Recently, with the help of data-driven approaches, many successes have been achieved and significantly reduced the gap for its application for downstream tasks in real life. However, most of these methods are either suffering from either strong stationary or linear dependencies assumptions~\cite{yu2016temporal, zhao2018correlation}, making them not suitable for real-life tasks with highly complicated and stochastic characterization in nature. Deep learning models also provide promising results and more flexibility for spatiotemporal dependencies modeling. Convolutional neural networks(CNN) have been used to model the regular 2-D structured traffic network in the transportation domain and have promising results in predicting crowd flow~\cite{zhang2017deep}. Graph neural network is another counterpart to model more complex spatial structure~\cite{ye2020build}. The graph convolutional network (GCN) proposed by Bruna et al. ~\cite{bruna2013spectral} first bridged the gap between the deep neural network and the graph spectral theory~\cite{bruna2013spectral} and then introduced in the traffic prediction problem by ~\cite{seo2018structured, gehring2017convolutional}. Graph attention network (GAT) is an alternative approach that can learn the attentional weights on each graph node and is broadly used in spatiotemporal forecasting problems~\cite{liu2019attention}. 

Though plenty of research exists in solving the traffic prediction problem, most rely on the assumption that the historical data of each location of interest in the prediction task are accessible. However, in real life, this assumption does not always hold. One way to solve this problem is to krige the unrecorded locations and then forecast the future time point. Some statical models, like using mean or zero to fill the unrecorded locations or regression-based models~\cite{ansley1984estimation}, matrix factorization~\cite{acar2011scalable}, are also used. However, the completed data used in training the prediction model will also bring in errors and weaken the prediction accuracy at locations with historical data. To avoid these problems, inspired by the inductive graph neural network~\cite{wu2021inductive}, we proposed a new framework that can directly execute prediction tasks at both locations with historical data and locations whose historical data are not accessible because of reasons like sensors not deployed at these points of interest.

\subsection{Inductive Graph Learning}
In recent years, graph neural network has attracted much attention due to their high expressiveness in capturing intricate relationships and dependencies of structured data and their flexibility to model irregular data~\cite{scarselli2008graph, duvenaud2015convolutional, kipf2016semi}. After the proven of both transductive and inductive representation capability~\cite{zeng2019graphsaint}, researchers have developed many inductive GNN methods in applications such as recommendation systems from learning embedding of each node in a fixed graph to learning to embed of node features that can be generated to unseen nodes\cite{zhang2019inductive}. For example, Zhang et al.~\cite{zhang2018gaan} train a GCN by masking a part of the observed user and item embedding and learning to reconstruct the masked ones. And Zeng~\cite{zeng2019graphsaint} trains GNNs for large graphs by sampling and learning from the subgraph. In inductive learning, learning node embedding through the structural properties of a node's neighbor properties instead of learning each node's embedding in a fixed graph is suitable. It can be easily adapted to the prediction problem at locations with and without historical data, as we sample subgraphs at locations with historical data and train inductive GNN models. Then the trained GNN model could treat locations without historical data as unseen nodes and embed the unseen nodes using the learned topological structure of these node's neighborhoods and the distribution of the node features in the neighborhood~\cite{zeng2019graphsaint}. Thus in this paper, we take the setting from the previous inductive graph neural network~\cite{wu2021inductive} and extend the task from spatiotemporal kriging into the prediction tasks with the ability to extend the perception to locations without historical data.

\subsection{Uncertainty Quantification}
Research in uncertainty quantification related to deep learning has become a thriving field in recent years. There are two mainstream uncertainty quantification methods, mostly beyond consideration. The first category is the Bayesian method which models the posterior distribution of the network parameters optimized with the data. Methods including dropout~\cite{gal2016dropout, gal2017concrete}, ensembling~\cite{lakshminarayanan2017simple}, and other approaches~\cite{blundell2015weight, hernandez2015probabilistic} are taken to quantify the variance and place the priors over network weights using variational inference~\cite{kingma2015variational}. A recent method, the evidential deep learning method, alternatively trains a deterministic model and places the uncertainty before the predictive distribution instead~\cite{amini2020deep} and can simultaneously detect OOD and adversarial data. The second category is the frequentist uncertainty quantification methods which focus on the robustness, not the variations in the data~\cite{wu2021quantifying}. Existing methods quantify uncertainty by taking the prediction intervals as the prediction objective~\cite{pearce2018high} or with bootstrapping using the influence function~\cite{alaa2020frequentist} for example. In this work, we investigate from both Bayesian and frequentist perspectives on our spatiotemporal forecasting problem. And we pay more attention to the evidential deep learning approach in the Bayesian framework as it can distinguish the OOD samples in the data, which can best help explain the high uncertainty predictions in prediction locations with no historical data.
\section{Preliminary}
In this section, we first introduce definitions and the original problem statement of traffic prediction problems following the previous works~\cite{li2017diffusion}.  Then we will extend the original traffic prediction problem to missing historical data locations and give a formal problem statement. All the notation is summarized in Table~\ref{tab: notations} for brevity.

\begin{table}[!ht]
\caption{Notations}
\label{tab: notations}
\resizebox{\linewidth}{!}{
\begin{tabular}{c|c}
\hline
\textbf{Notations} & \textbf{Descriptions} \\ \hline
    $N$                 &         Number of locations of interest in the network              \\ \hline
    $N_o, N_e$                 &         Number of observable/missing locations               \\ \hline
    \multirow{2}{*}{$\mathcal{V}$}                 &     Nodes representing the locations of interest \\ & in the traffic network                \\ \hline
    \multirow{2}{*}{$\mathcal{A}$}                &         Adjacent matrix representing the connectivity\\ & of  the traffic network              \\ \hline
    \multirow{2}{*}{$\mathcal{A}_{o}$}                &         Adjacent matrix representing the connectivity \\ & of observable locations              \\ \hline    
    $\mathcal{G(\cdot , \cdot)}$               &    Graph representation of the network                   \\  \hline
    $\mathcal{V}_o , \mathcal{V}_e$            &    Nodes at observable/missing locations                   \\ \hline
    \multirow{2}{*}{$\mathbf{X}^{(t)}$}       &     Node features at observable locations and
    \\ & 0 at missing locations at time step $t$               \\ \hline
    $\mathbf{X}_o^{(t)}$          &    Node features at observable locations at time step $t$ \\  \hline
    \multirow{2}{*}{$\mathbf{\hat{X}}_o^{(t)}, \mathbf{\hat{X}}_e^{(t)}$}          &    Predicted node features at observable/missing \\ & locations  at time step $t$ \\ \hline
    $h(\cdot)$ & The function to predict future node features at $\mathcal{V}_{o}$  \\ \hline
    $f(\cdot)$ & {\begin{tabular}[c]{@{}c@{}}The function to predict future node features at  \\  all locations of interest $\mathcal{V}_{o}$ and $\mathcal{V}_{e}$ \end{tabular}} \\ \hline  
\end{tabular}
}
\end{table}

Suppose we have predefined $N$ locations of interest in the traffic network which are deemed critical and have a strong desire to acquire timely information. We present this traffic network as a directed graph $\mathcal{G} = (\mathcal{V}, \mathcal{A})$, where $\mathcal{V}$ is a set of nodes and the cardinality of the nodes $|\mathcal{V}|$ is $N$. $\mathcal{A}$ is the adjacent matrix describing the connectivity of the network $\mathcal{G}$ and $\mathcal{A} \in \mathbb{R}^{N \times N}$. 

In real life, due to the accessibility of sensors at some given locations, $N_o$ out of $N$ locations, which are a subset $\mathcal{V}_{o} \subseteq \mathcal{V}$ (also known as the observed locations) represented on $\mathcal{G}$ have sensors deployed and historical data $\mathbf{X}_{o} \in \mathbb{R}^{N_o \times P}$ are observable, where $P$ is the number of features of each node. $\mathbf{X}_{o}^{(t)}$ represents the traffic state observed at time t and location $\mathcal{V}_{o}$. And the sub-graph of these observed locations is described as $\mathcal{G}_{o} = (\mathcal{V}_{o}, \mathcal{A}_{o})$, where $\mathcal{A}_{o}$ is the adjacent matrix describing the connectivity between observed locations. For the rest of the locations of interest with no historical records at $\mathcal{V}_{e} \subseteq \mathcal{V}$ (also known as the missing location), we denote historical data at time t as $\mathbf{X}_{e}^{(t)} \in \mathbb{R}^{N_e \times P}$, where $N_e$ is the number of missing locations.

\begin{problem}[\textbf{Traffic Prediction Problem}]

In the traffic prediction problem, at time step $t$, we have T length of historical traffic state $\mathbf{X}_{o}^{(t-T+1)}, \mathbf{X}_{o}^{(t-T+2)},\cdots, \mathbf{X}_{o}^{(t)}$ at locations $\mathcal{V}_{o}$ and the network structure $\mathcal{G} $, the goal is to learn a function $h{(\cdot)}$ that takes the historical traffic states and graph structure and predict the traffic states at locations $\mathcal{V}_{o}$ at the future time $t + \Delta t$:
\begin{equation}
    [\mathbf{X}_{o}^{(t-T+1)}, \mathbf{X}_{o}^{(t-T+2)}, \cdots, \mathbf{X}_{o}^{(t)}, \mathcal{G}] \xrightarrow{h(\cdot)} \mathbf{\hat{X}}^{(t+\Delta t)}_{o}
\end{equation}
\end{problem}

In real-life scenarios, due to budget or physical restrictions, only part of the locations are observable and have accessibility to historical records. However, the information in other missing locations is also critical and of interest; for example, the predicted future speed could help plan routines for vehicles or to control traffic signals for better traffic regulation; we need to predict traffic state at both observable locations $\mathcal{V}_{o}$ and extend the perception to missing locations $\mathcal{V}_{e}$. Based on the differences mentioned above, we extend the original problem to the traffic prediction with missing locations problem and formally define it as follows:

\begin{problem}[\textbf{Traffic Prediction with Missing Locations Problem}]
\label{problem: 2}
In the traffic prediction with missing locations problem, following the original setting~\cite{li2017diffusion}, we have the historical observations at locations $\mathcal{V}_{o}$ and the network structure. In the new problem, the goal is to learn a function $f(\cdot)$ that takes the historical traffic state at locations $\mathcal{V}_{o}$ and traffic network graph structure $\mathcal{G}$ and predicts the future traffic state at both observed locations $\mathcal{V}_{o}$ and missing locations $\mathcal{V}_{e}$ at the future time points $t+\Delta t$:
\begin{equation}
\label{p: new problem}
    [\mathbf{X}_{o}^{(t-T+1)}, \mathbf{X}_{o}^{(t-T+2)}, \cdots, \mathbf{X}_{o}^{(t)}, \mathcal{G}] \xrightarrow{f(\cdot)} [\mathbf{\hat{X}}^{(t+\Delta t)}_{o}, \mathbf{\hat{X}}^{(t+\Delta t)}_{e}]
\end{equation}

\end{problem}

\section{Method}
While traffic prediction is a well-studied field, making prediction extendable to locations with no historical data is still challenging due to the lack of prior knowledge and the complex spatial dependencies on the traffic network. To overcome these problems, we adopt the inductive graph neural network, which is used initially in kriging tasks, and extend it to traffic prediction. At the same time, we integrate the uncertainty quantification component into our prediction model to evaluate the confidence of the prediction in the absence of ground truth data. The proposed uncertainty-aware inductive neural network (UIGNN) framework can predict traffic state at observable locations with historical data and extend prediction perception to missing locations with no historical data while quantifying the uncertainty of predictions. In the rest of this section, we will briefly introduce the UIGNN framework. 

\subsection{Traffic Prediction with Missing Locations}
In the setting defined by Problem~\ref{problem: 2}, the model $f(\cdot)$ should be applicable to predict observed locations $\mathcal{V}_{o}$ and missing locations $\mathcal{V}_{e}$. Most current methods in conventional traffic prediction problems cannot be directly used in this scenario because of the lack of training data at the missing locations. To stress this problem, inspired by graph sampling-based inductive learning methods~\cite{hamilton2017inductive} and its recent application in spatiotemporal kriging problem~\cite{wu2021inductive}, we take an inductive learning method that models the traffic network with graph neural network and take a sampling-based method to learn the representation of nodes in the sampled graph and generalize those representations to unseen nodes and graphs. The inductive graph neural network (inductive GNN) can learn the topological structure and distribution of the node features in the neighborhood and is thus applicable to graphs both with and without node features w.r.t observable locations and missing locations, respectively. In this work, we follow the past IGNNK~\cite{wu2021inductive} and extend the task from kriging node features 
 $\mathbf{\hat{X}}_{e}^{(t)} = g(\mathbf{X}_{o}^{(t-T+1)}, \mathbf{X}_{o}^{(t-T+2)}, \cdots, \mathbf{X}_{o}^{(t)}, \mathcal{G})$, where $g(\cdot)$ is the learned function to kriging the missing locations, into the traffic prediction with missing location problem. Below are the details of how we take the inductive learning method and train the model to learn the node representation only on observable locations and generalize the learned node representation to predict future traffic states on observable and missing locations the model has not seen.

\subsubsection{\textbf{Inductive Graph Neural Network}}
To overcome the lack of historical data and learn a node representation generalized to $\mathcal{V}_{e}$, we follow the implementation of IGNNK to train an inductive graph neural network by sampling sub-graph node features at $\mathcal{V}_o$. Following the inductive learning approach, there is an assumption that the observable locations represent the overall population distribution, and the message-passing mechanism could be generated to unseen locations $\mathcal{V}_{e}$. 

Considering our task is to predict traffic network state with only partial locations observable, during training, our inductive model should also be able to learn node representation with part of node features available such that the learned model could predict all locations $\mathcal{V}$ with only historical data at observed locations $\mathcal{V}_{o}$. Thus we take a sampling strategy to train our model on $\mathcal{G}_{s} \subseteq \mathcal{G}_{o}$ instead of $\mathcal{G}_{o}$.

\subsubsection{\textbf{Random Sample of Sub-graph}}

To prepare the sample for training GNN, we first generate random integers $n_s \leq n_o$, which is the number of sampled nodes $\mathcal{V}_{s} \subseteq \mathcal{V}_o$. And next, we split sampled nodes $\mathcal{V}_{s}$ into two groups $\mathcal{V}_r$ and $\mathcal{V}_m$ with carnality of $|\mathcal{V}_{r}|=n_{r}, |\mathcal{V}_{m}|=n_{m}, n_{r} + n_{m} = n_{s}$. And we reserve the observed historical data at $\mathcal{V}_{r}$ and mask the observed historical data at $\mathcal{V}_{m}$ to make node features partially accessible. Thus we have the set of sampled nodes, reserved nodes, and masked nodes in the network following the relationship: $\mathcal{V}_s = \mathcal{V}_r \cup \mathcal{V}_m$. To make annotation simple, we use $\mathcal{V} \subset \mathcal{G}$ with slight abuse to represent the node $\mathcal{V}$ itself and also its index in the graph $\mathcal{G}$. For example, $i \in \mathcal{V}_{r} \subset \mathcal{G}_{s}$ means the set of index belongs to observed nodes in the sampled graph $\mathcal{G}_{s}$. 

Then we generate the mask matrix $\mathcal{M}_{s} \in \mathbb{R}^{n_s \times P}$ with 
\begin{equation}\
    \mathcal{M}_{s}[i,: ] = 
    \begin{cases}
        1, & \text{if } i \in \mathcal{V}_r \subset \mathcal{G}_{s}, \\
        0, & \text{otherwise} 
    \end{cases}
\end{equation}
and construct adjacent matrix $\mathcal{A}_s$ of sub-graph $\mathcal{G}_s$ by selecting entries at $\mathcal{A}_{i,j}$ of $\mathcal{A}$, where $i,j \in \mathcal{V}_s \subset \mathcal{G}$. And the sampled graph could be represent as $\mathcal{G}_{s} = (\mathcal{V}_{s}, \mathcal{A}_{s})$

We finally randomly sample time step $t$ and pick $\mathbf{X}_s^{(t-T+1:t)}$ at $\mathcal{V}_{s}$ from $t-T+1$ to $t$ time interval, where $T$ is the length of the historical data taken by the model as input node features and  $\mathbf{X}_s^{(t+\Delta t)}$ as the target. 

We iterate for $I$ iterations, and in each iteration, we sample sub-graph $\mathcal{G}_{s} \text{ for } S$ times and train the GNN model $f(\cdot)$ with the samples described  above. The detail is shown in Algorithm~\ref{algorithm: subgraph}. Then we apply the trained $f(\cdot)$ in the prediction task on the graph $\mathcal{G}$.


\begin{algorithm}[htb]

\DontPrintSemicolon
\caption{Sub-graph sampling and random mask for GNN training}
\label{algorithm: subgraph}

\KwIn{Historical data $\mathbf{X}$ of the traffic network, graph structure of traffic network $\mathcal{G}(\mathcal{V}, \mathcal{A})$, initialized GNN model $f(\cdot)$}
\KwOut{trained GNN model $f(\cdot)$}
\For {i = 1,2, \dots, I} {
    create sampl mask list $\mathcal{M}^{l}$ = [], adjacent matrix list $\mathcal{A}^l$ = [], feature list $\mathbf{X}^{l}$ = [], target list $\mathbf{Y}^l$ = [] \\
    \For {s = 1, 2 \dots, S} {
        Generate number $n_r, n_m$ \\
        Sample $n_r$ of nodes $\mathcal{V}_r \subset \mathcal{V}_o$, $n_m$ of nodes $\mathcal{V}_m \subset \mathcal{V}_o \setminus \mathcal{V}_r$ and $\mathcal{V}_{s} = \mathcal{V}_{r} \cup \mathcal{V}_{m}$\\
        Construct adjacent matrix $\mathcal{A}_s$ by selecting entries at $\mathcal{A}_{i,j}$, where $i,j \in \mathcal{V}_s \subset \mathcal{G}$, and construct sub-graph $\mathcal{G}_{s} = (\mathcal{V}_{s}, \mathcal{A}_{s})$. $\mathcal{A}^l$.append($\mathcal{A}_s$)\\        
        Generate mask matrix $\mathcal{M}_s$ with $1$ at row $i \in \mathcal{V}_r \in \mathcal{G}_{s}$ and $0$ at row $j \in \mathcal{V}_m \subset \mathcal{G}_{s}$, $\mathcal{M}^{l}$.append($\mathcal{M}_s$) \\

        Random sample time step $t$ and obtain node features $\mathbf{X}_s^{(t-T+1:t)}$ at $\mathcal{V}_s$ for $T$ historical length and target $\mathbf{X}_s^{(t+\Delta t)}$,  $\mathbf{X}^{l}$.append($\mathbf{X}_s^{(t-T+1:t)}$), $\mathbf{Y}^l$.append($\mathbf{X}_s^{(t+\Delta t)}$)\\
    }
    Train GNN model $f(\cdot)$ with sampled data $\mathcal{M}^{l}$, $\mathcal{A}^l$, $\mathbf{X}^{l}$, $\mathbf{Y}^l$.
}
\end{algorithm} 

\subsubsection{\textbf{Diffusion Graph Convolutional Network}}
To capture the stochastic nature of spatiotemporal dependencies in the directional network, we follow the past work and adopt Diffusion Graph Convolutional Network\cite{atwood2016diffusion} (DGCNs) as the basic building block for the GNN architecture. The DGCNs model the message passing on the GNN as a diffusion process that explicitly captures the stochastic nature of the transition dynamics and is theoretically and experimentally justified applied to spatiotemporal forecasting problems. The basic building block could be represented as:
\begin{equation}
\label{eq: dgcn}
    H_{l+1} = \sum^{K}_{k=1}T_k(\bar{\mathcal{A}}_f) H_l \Theta^{k}_{b,l} + T_k(\bar{\mathcal{A}}_b) H_l \Theta^k_{f,l}
\end{equation}
where $\bar{\mathcal{A}}_b$ is $\mathcal{A}$ normalized across each row and $\bar{\mathcal{A}}_b$ = $\bar{\mathcal{A}}_f.\text{transpose()}$. The two matrices represent the forward and backward transition matrices, respectively. We have $T_k(\bar{\mathcal{A}}) = 2\bar{\mathcal{A}}T_{k-1}(\bar{\mathcal{A}}) - T_{k-2}(\bar{\mathcal{A}})$ and $T_0(\bar{\mathcal{A}}) = I, T_1(\bar{\mathcal{A}}) = \bar{\mathcal{A}}$ for the first and second order, where k is the order of the Chebyshev polynomial used in the DGCN block. And the $\Theta^k_{f,l}$ and $\Theta^k_{b,l}$ are learnable parameters of GNN model. The input layer in our traffic forecasting problem is 
\begin{equation}
    H_0 = \mathbf{X}_{s} \otimes \mathcal{M}_s
\end{equation}
, where $\otimes$ is the Hadamard product. Since $H_{0}[j, :]$, where$j \in \mathcal{V}_s \subset \mathcal{G}_{s}$ is all $0$ after masked by $\mathcal{M}_s$, we follow~\cite{wu2021inductive} and define the second layer as:
\begin{equation}
    H_{2} = \sigma(\sum^{K}_{k=1}T_k(\bar{\mathcal{A}}_f) H_1 \Theta^{k}_{b,1} + T_k(\bar{\mathcal{A}}_b) H_1 \Theta^k_{f,1}) + H_{1}
\end{equation}
where $\sigma(\cdot)$ is the nonlinear activation function for the DGCN layers. We add the $H_{1}$ to the $H_{2}$ since it contains the information at the $\mathcal{V}_m$ locations with no sensors. The prediction target is the output of the last layer from DGCN, changing into the right shape through a fully connected layer. During training, the forward and backward transition adjacent matrix is $\bar{\mathcal{A}}_s$ calculated from $\mathcal{A}_s$ for learning the graph representation and parameter $\{\Theta^{k}_{f,l}, \Theta^{k}_{b,l} \}^{\{1, 2, \cdots, K\}}_{\{1,2, \cdots, L\}}$, where $K$ is the order of Chebyshev polynomial and $L$ is the number of layers of GNN model. 

Since our task is to predict future traffic state on observable and missing locations on $\mathcal{G}$, the training procedure on the sampled network $\mathcal{G}_s$ should be able to predict reserved locations observable after masking and recover locations masked from observation such that the message passing mechanism could be generalized to all nodes. As a result, the learned inductive GNN model trained on $\mathcal{G}_s$ could be generalized to network $\mathcal{G}$. To achieve this, we design the prediction loss function as follows:

\begin{equation}
    J_{pre} = \sum_{s \in S} Loss(X_{s}^{t+\Delta t}, \text{MLP}(f(\mathbf{X}_{s}^{(t-T+1:t)} \otimes \mathcal{M}_s, \mathcal{A}_s)))
\end{equation}
where $\text{MLP}$ means the fully connected layer for final outputs.

To keep more structure and feature information, we also minimize the difference between recovered data and initial model inputs by defining a recovery loss as follows:

\begin{equation}
    J_{rec} =  \sum_{s \in S} Loss(f(\mathbf{X}_{s}^{(t-T+1:t)} \otimes \mathcal{M}_s, \mathcal{A}_s),H_{0})
\end{equation}
And we can define our total loss as follows:
\begin{equation}
    J_{total} = J_{pre}+\alpha J_{rec}
\end{equation}
where $\alpha$ is a hyperparameter for balancing the two loss functions. 





\subsection{Uncertainty Aware Traffic Forecasting}
In this section, we will introduce the limitations of most current models based on point estimation~\cite{wu2021quantifying} in our Problem~\ref{problem: 2} and propose a new method that can predict the future traffic state and quantify the uncertainty at each location in the traffic network and show how the down-stream tasks can benefit from it. 

\subsubsection{Model Uncertainty on Sensor Network}
The past works in traffic prediction problems usually use point estimation rather than probabilistic prediction with built-in uncertainty. However, accurate point estimations often require evaluation metrics like root mean square error (RMSE), which is not retrievable at missing locations with no historical data. As a result, model performance can only be evaluated at observable locations, and we will have no sense of the model performance at missing locations. And the resulting predictions will not be accredited in the downstream tasks that need assessing potential risks and involve decision-making.

To overcome the limitations of point estimation and asses sable at missing locations, we need to quantify the uncertainty at different locations of traffic networks, especially at the extended missing ones. Specifically, we would like to output the uncertainty value along with the prediction value at each location:

\begin{align}
    [\mathbf{X}_{o}^{(t-T+1)}, \mathbf{X}_{o}^{(t-T+2)}, \cdots, \mathbf{X}_{o}^{(t)}, \mathcal{G}] \xrightarrow{f(\cdot)} & [\mathbf{\hat{X}}^{(t+\Delta t)}_{o}, \mathbf{\hat{X}}^{(t+\Delta t)}_{e}], \notag \\ 
    & [\mathbf{U}^{(t+\Delta t)}_{o}, \mathbf{U}^{(t+\Delta t)}_{e}]
\end{align}
where the $\mathbf{U}^{(t+\Delta t)}_{o}, \mathbf{U}^{(t+\Delta t)}_{e}$ are uncertainty at observable and missing locations respectively.

Any uncertainty quantification method applies to our proposed framework as long as it can quantify the uncertainty of the deep neural network in regression tasks such as evidential deep regression~\cite{amini2020deep} (EDL), Dropout~\cite{gal2016dropout} (Dropout), Quantile Regression~\cite{koenker2001quantile} (QR), etc. In this paper, we use EDL to quantify model uncertainty and investigate the uncertainty at both observable and missing locations, since it is most compatible with our Problem~\ref{p: new problem} setting.

Intuitively, the inductive graph neural network enables our model to predict future traffic states in both observable and missing locations; However, due to the absence of node features at missing locations, the information may be lost and thus bring in additional data uncertainty compared to observable locations with node features. Besides, due to the topological structure and distribution of node features in the neighborhood in the traffic network graph, some of the nodes' representation may not be well-represented during the training process and, as a result, have higher model uncertainty compared to other locations. And finally, the out-of-distribution nodes at missing locations will have high prediction uncertainty. 

To quantify the uncertainty at each location, we incorporate the uncertainty quantification layer with the output layer and output the negative log-likelihood along with the predicted traffic states.

\subsection{Computational Complexity Analysis}

To demonstrate the efficiency of our proposed model, we discuss the advantage of taking diffusion graph neural network (DGCN) as our model backbone here. In general, computing a convolution on the graph is expensive. However, since $\mathcal{G}$ is sparse due to the nature of traffic network connection and use of Gaussian thresholded kernel, Equation~\ref{eq: dgcn} could be computed efficiently using $\mathcal{O}(K)$ recursive sparse-dense matrix multiplication~\cite{li2017diffusion}. The overall total time complexity is $\mathcal{O}(K|\mathcal{E}|) 	\ll \mathcal{O}(|\mathcal{V}^{2}|)$, which makes our model able to be deployed on real-time hardware with a limited computational resource. 
\section{Experimental evaluation}
To evaluate our model's performance in the Problem~\ref{problem: 2} and justify that our new problem setting holds significant relevance and could serve as guidance for different downstream tasks, we focus on the following research questions: 

\textbf{RQ1:} Comparing to other approaches applicable in our problem setting, how does our model UIGNN perform? 

\textbf{RQ2:} Does our model veritably reflect the prediction uncertainty at different sensor locations?

\textbf{RQ3:} Does the model trained in our new problem setting applicable and guide the downstream task?

\subsection{Dataset}
We conduct our experiments on three real-world, large-scale traffic datasets. \textbf{METR-LA: }This dataset contains traffic information from loop detectors located in the Los Angeles highway road net. We follow the settings of IGNNK~\cite{wu2021inductive} and select 207 sensors with a 6-month record in total to construct a spatiotemporal graph.  \textbf{PeMS-BAY: } It's a dataset similar to METR-LA, collected from highway networks with 325 sensors. Different from METR-LA, the weighted adjacency matrix is not constructed by latitude and longitude for the sensors listed directly, but by matching absolute postmile markers to find connectivity information. \textbf{SeData:} This dataset is collected from 323 locations with loop detectors on highway road net, following the same format as METR-LA. Its adjacency matrix is a simple binary matrix representing whether two nodes are connected or not.

In these datasets, we choose 70$\%$ of the data for training and the rest for validation and testing. During the sampling period, we randomly select a subset of sensors as unknown ones removed from the full set(the number changes when using different datasets). This removed set is defined as missing locations ($\mathcal{V}_{e}$) without historical data available and cannot be used as the target in training. And the remained set is denoted as the observable locations ($\mathcal{V}_{o}$) with historical data available during training and inference.

The adjacent matrix is computed with the thresholded Gaussian kernel~\cite{shuman2013emerging}, paired wisely from the distances along the road network between the two locations to construct the traffic sensor network graph. It is related to the direction of the road network. Each entry $A_{i,j} = exp(-\frac{dist(v_i, v_j)^{2}}{\sigma^{2}})$ and trucked to be $0$ if $dist(v_i, v_j) \leq \kappa$, where $\sigma$ is the standard deviation of distances and $\kappa$ is the threshold.


\subsection{Experiment Setup}
\subsubsection{\textbf{Baselines}}
Since our problem predicts future traffic state with historical data at partial locations, this is of major difference to past traffic prediction problems, and most past models cannot be directly adopted in this scenario. Thus, we take a two-step approach first, predict the historical traffic state at locations without sensors (missing locations) and then forecast the future traffic state based on the completed historical data. To make the comparison fair, we use the diffusion graph convolution neural network (DGCN) as the building block, the same as our UIGNN model. We pick 1) \textbf{Kriging + DGCN}: kriging is a spatial interpolation model~\cite{cressie2015statistics} widely used in the geostatistical field to recover data at missing locations and then use DGCN to forecast future traffic state at time $t + \Delta t$. We use the ordinary Kriging method, and the correlation is solely dependent on the network road distances instead of the spatial distance since it is on a road network structure.  2) \textbf{KNN + DGCN:} K spatially nearest neighbors of the missing locations are aggregated and averaged to estimate their traffic states. And the prediction step is the same as Kriging + DGCN. 3) \textbf{ST-MVL + DGCN:} ST-MVL~\cite{yi2016st} is a spatiotemporal multiview-based learning algorithm to recover missing data of geo-sensory time series data. In our setting, since missing locations, the historical data is unavailable; we only use spatial view to complete data. The prediction step is the same as the previous methods. 4) \textbf{IGNNK + DGCN:} We first train an IGNNK model to complete data at missing locations following the original work procedure~\cite{wu2021inductive}. The prediction step is the same as previous methods. Notice matrix factorization-based methods do not apply to our problem since they need future traffic states at observable locations to recover the missing locations at the test dataset. Thus it is not fair to compare. 


\subsubsection{\textbf{Hyperparameters}} The dimension of the hidden layers of our model is set to 100. We implement all our batch sizes in our experiments set to 4. We use the Adam optimizer, set the learning rate to 1e-4, and train the UIGNN model for 750 epochs. We choose time slice h=24 for METR-LA, PeMS-bay, and SeData.

\subsubsection{\textbf{Metrics}}
To compare our model and other baseline model performance, we use mean RMSE at 30 mins for all three traffic speed datasets and split the locations into two groups: observed and missing. To directly show our prediction and uncertainty quantification results, we report the root mean square error (RMSE) and negative log-likelihood (NLL) loss of all sensors in our experiments and split them into two groups: observable and missing locations. We report 15-, 30- and 60-min results for all three datasets to investigate their prediction performance at different time scales. 

\subsection{Implementation of Downstream Task}
To answer \textbf{RQ3}, we evaluate our model for the active sensing downstream task in a traffic sensor network~\cite{krause2008near, guestrin2005near}. This task involved deploying sensors at a subset of locations and gradually adding sensors based on the model's output at each step. Due to budget constraints and communication costs, this active sensing scenario is crucial in real-life situations, and the current traffic prediction problem has not yet been stressed.

The experiment is conducted on the METR-LA dataset and predicts the traffic state for the upcoming 30 minutes. During the initial phase, we randomly selected 50 sample locations from 207 locations in the traffic sensor network. These selected locations $\mathcal{V}_{o}$ among all locations $\mathcal{V}$ are designated as "deployed with sensors," indicating that historical data was observable. We proceeded to train our model using the available data from these 50 locations. In the active sensing phase, new sensors are available with a fixed budget (10 sensors are deployed at each step). At each stage, we sample ten new locations from $\mathcal{V} \setminus \mathcal{V}_{o} $ based on the highest uncertainty indicated by our model. We then made the historical data available for these newly selected locations. Subsequently, we retrained our model using both the previously observable locations and the newly acquired data.

\section{Experimental results}

\subsection{Overall Performance of UIGNN (\textbf{RQ1})}
\begin{table*}[!t]
\caption{Overall performance of UGINN and other baseline methods on 30 mins prediction task.For MAE and RMSE, the lower, the better. For $R^2$, the higher, the better. The \underline{\textbf{Best}} and \underline{ second best} performances are highlighted. `'-' means the performance cannot converge.}
\label{tab: baseline}
\centering
\resizebox{0.85\linewidth}{!}{
\begin{tabular}{c||c|c|ccccc}
\toprule\toprule
\multirow{2}{*}{\textbf{Dataset}} &
  \multirow{2}{*}{\textbf{Location}} &
  \multirow{2}{*}{\textbf{Metric}} &
  \multicolumn{5}{c}{\textbf{Method}} \\
 &
   &
   &
  \multicolumn{1}{c}{\textbf{UIGNN}} &
  \multicolumn{1}{c}{\textbf{Kriging+DGCN}} &
  \multicolumn{1}{c}{\textbf{KNN+DGCN}} &
  \multicolumn{1}{c}{\textbf{ST-MVL+DGCN}} &
  \textbf{IGNNK+DGCN} \\ 

  \midrule
\multirow{6}{*}{\rotatebox{90}{\textbf{METR-LA}}} &
  \multirow{3}{*}{\textbf{Observable}} &
  \textbf{MAE} &
  \multicolumn{1}{c}{\underline{\textbf{6.5677}}} &
  \multicolumn{1}{c}{7.5525} &
  \multicolumn{1}{c}{\underline{7.2840}} & 
  \multicolumn{1}{c}{7.5547} &
  7.2916\\ 
 &
   &
  \textbf{RMSE} &
  \multicolumn{1}{c}{\underline{\textbf{3.9850}}} &
  \multicolumn{1}{c}{4.1465} &
  \multicolumn{1}{c}{\underline{4.0180}} &
  \multicolumn{1}{c}{4.3918} & 4.0263
   \\
 &
   &
  \textbf{R$^2$} &
  \multicolumn{1}{c}{\underline{\textbf{0.9141}}} &
  \multicolumn{1}{c}{0.8917} &
  \multicolumn{1}{c}{\underline{0.8992}} &
  \multicolumn{1}{c}{0.8916} & 0.8991
   \\ \cline{2-8} 

 &
  \multirow{3}{*}{\textbf{Missing}} &
  \textbf{MAE} &
  \multicolumn{1}{c}{\underline{\textbf{10.5004}}} &
  \multicolumn{1}{c}{16.7824} &
  \multicolumn{1}{c}{14.4969} &
  \multicolumn{1}{c}{11.8830} & \underline{10.8898}
   \\
 &
   &
  \textbf{RMSE} &
  \multicolumn{1}{c}{\underline{\textbf{7.3052}}} &
  \multicolumn{1}{c}{12.0907} & 
  \multicolumn{1}{c}{10.5916} &
  \multicolumn{1}{c}{8.0088} & \underline{7.7083}
   \\
 &
   &
  \textbf{R$^2$} &
  \multicolumn{1}{c}{\underline{\textbf{0.7997}}} &
  \multicolumn{1}{c}{0.3950} &
  \multicolumn{1}{c}{0.6214} &
  \multicolumn{1}{c}{0.7456} & \underline{0.7865}
   \\ 
   \midrule
\multirow{6}{*}{\rotatebox{90}{\textbf{PeMS-BAY}}} &
  \multirow{3}{*}{\textbf{Observable}} &
  \textbf{MAE} &
  \multicolumn{1}{c}{\underline{4.8879}} &
  \multicolumn{1}{c}{\textbf{---}} &
  \multicolumn{1}{c}{4.4959} &
  \multicolumn{1}{c}{4.5611} & 
  \underline{\textbf{4.4065}} \\ 
 &
   &
  \textbf{RMSE} &
  \multicolumn{1}{c}{\underline{\textbf{1.9735}}} &
  \multicolumn{1}{c}{\textbf{---}} &
  \multicolumn{1}{c}{2.1616} &
  \multicolumn{1}{c}{2.2434} & \underline{2.1282}
   \\ 
 &
   &
  \textbf{R$^2$} &
  \multicolumn{1}{c}{\underline{0.8062}} &
  \multicolumn{1}{c}{\textbf{---}} &
  \multicolumn{1}{c}{0.8006} &
  \multicolumn{1}{c}{0.7948} & \underline{\textbf{0.8085}}
   \\ \cline{2-8} 
 &
  \multirow{3}{*}{\textbf{Missing}} &
  \textbf{MAE} &
  \multicolumn{1}{c}{\underline{\textbf{6.9762}}} &
  \multicolumn{1}{c}{\textbf{---}} &
  \multicolumn{1}{c}{7.9449} &
  \multicolumn{1}{c}{17.6647} & \underline{6.9922}
   \\ 
 &
   &
  \textbf{RMSE} &
  \multicolumn{1}{c}{\underline{\textbf{4.1054}}} &
  \multicolumn{1}{c}{\textbf{---}} &
  \multicolumn{1}{c}{5.0034} & 
  \multicolumn{1}{c}{9.8226} & \underline{4.1738}
   \\ 
 &
   &
  \textbf{R$^2$} &
  \multicolumn{1}{c}{\underline{0.4518}} &
  \multicolumn{1}{c}{\textbf{---}} &
  \multicolumn{1}{c}{0.3138} &
  \multicolumn{1}{c}{-4.5843} & \underline{\textbf{0.4688}}
   \\
   \midrule
\multirow{6}{*}{\rotatebox{90}{\textbf{SeData}}} &
    \multirow{3}{*}{\textbf{Observable}} &
  \textbf{MAE} &
  \multicolumn{1}{c}{\underline{\textbf{6.4863}}} &
  \multicolumn{1}{c}{6.6840} &
  \multicolumn{1}{c}{6.7028} &
  \multicolumn{1}{c}{6.9860} &
  \multicolumn{1}{c}{\underline{6.5252}} \\ 
 & 
   &
  \textbf{RMSE} &
  \multicolumn{1}{c}{\underline{3.8114}} &
  \multicolumn{1}{c}{3.9150} &
  \multicolumn{1}{c}{3.8350} &
  \multicolumn{1}{c}{4.2137} &  \underline{\textbf{3.7636}}
   \\ 
 &
   &
  \textbf{R$^2$} &
  \multicolumn{1}{c}{\underline{\textbf{0.7299}}} &
  \multicolumn{1}{c}{0.7132} &
  \multicolumn{1}{c}{0.7116} &
  \multicolumn{1}{c}{0.6867} & \underline{0.7281}
   \\ \cline{2-8} 
 &
  \multirow{3}{*}{\textbf{Missing}} &
  \textbf{MAE} &
  \multicolumn{1}{c}{\underline{8.1671}} &
  \multicolumn{1}{c}{10.6495} &
  \multicolumn{1}{c}{12.3653} &
  \multicolumn{1}{c}{13.0136} & \underline{\textbf{7.8984}}
   \\ 
 &
   &
  \textbf{RMSE} &
  \multicolumn{1}{c}{\underline{\textbf{4.9139}}} &
  \multicolumn{1}{c}{7.0653} &
  \multicolumn{1}{c}{9.9310} &
  \multicolumn{1}{c}{7.6135} & \underline{4.9779}
   \\
 &
   &
  \textbf{R$^2$} &
  \multicolumn{1}{c}{\underline{\textbf{0.6103}}} &
  \multicolumn{1}{c}{0.2693} &
  \multicolumn{1}{c}{0.2567} &
  \multicolumn{1}{c}{0.0911} & \underline{0.6002}
   \\
   \bottomrule
\end{tabular}
}
\end{table*}

To evaluate traffic prediction performance, we separately report the results of our method and four other two-step approaches on three traffic speed datasets in terms of root mean square error (RMSE), mean absolute error (MAE). We additionally measure the coefficient of determination ($\text{R}^2$) to describe the proportion of the variation in prediction errors over historical mean values.

The result in Table~\ref{tab: baseline} shows that our UIGNN achieves the best or second best performance in all three datasets compared to all two-step-based (first impute then predict) approaches on both observable and unobservable locations. On unobservable locations, our result shows UIGNN can accurately forecast future traffic states at locations without any historical records in the sensor network. While on observable locations, it also achieves promising results without hurting its accuracy on the traditional traffic prediction problem. We also find the performance of two-step approaches largely relies on the performance of the imputation model in the first step. While on missing locations, since the imputed input is not accurate and thus brings in errors in node representation learning hence weakens the performance at these locations. Since the IGNNK + DGCN method is a learning-based model and achieves higher accuracy in the data imputation step. As a result, it outperforms other baseline models. And kriging+DGCN method cannot converge since the imputation error on missing locations is too large.

\subsection{Uncertainty on Sensor Network (\textbf{RQ2})}
\begin{figure*}[!htp]
\centering
  \begin{tabular}{cc}
  \includegraphics[width=0.4\linewidth]{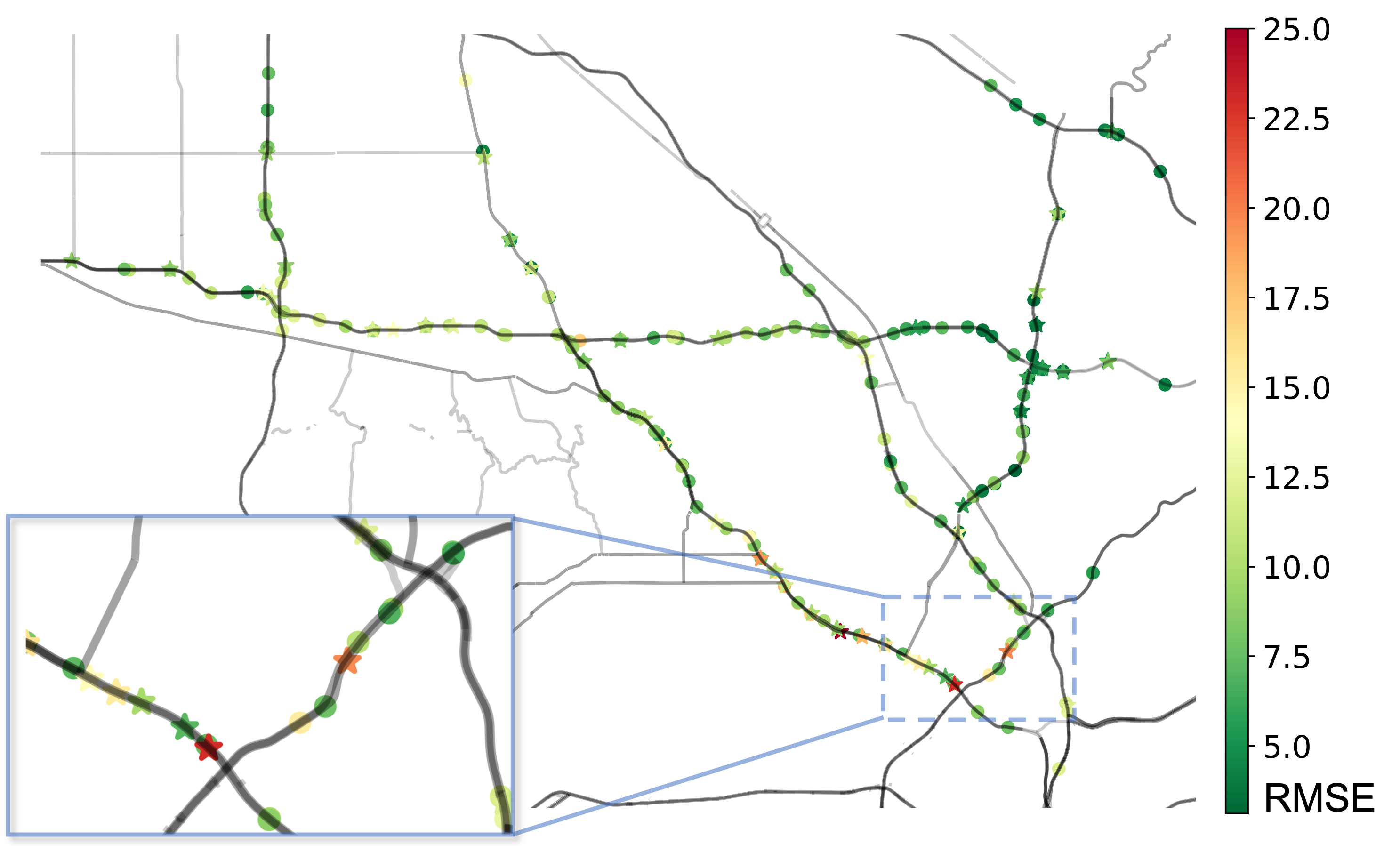}
  &

  \includegraphics[width=0.4\linewidth]{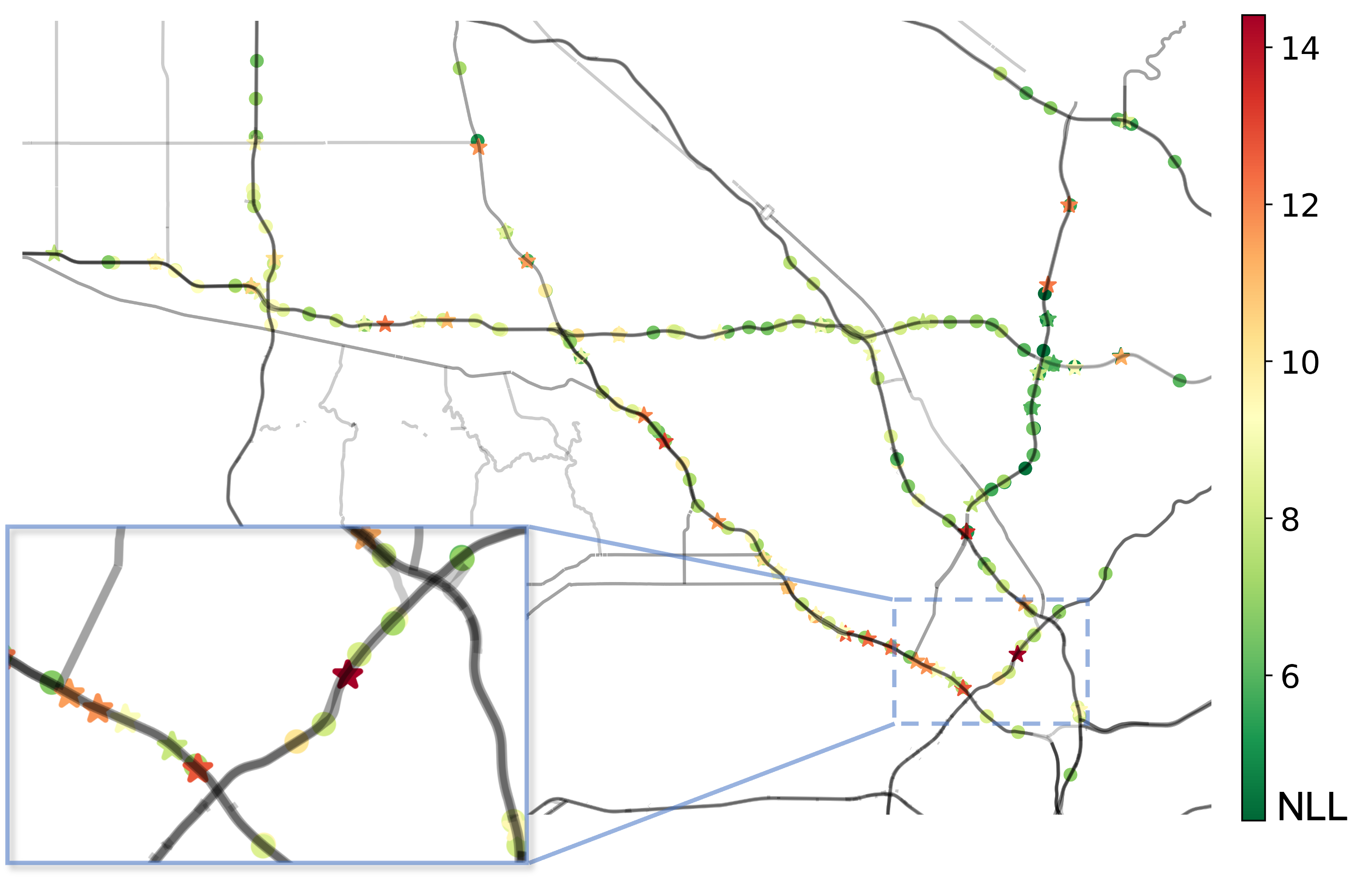}
  \\
  (a) RMSE on METR-LA traffic network & (b) NLL on METR-LA traffic network
  \vspace{1em}
  \\
   \includegraphics[width=0.4\linewidth]{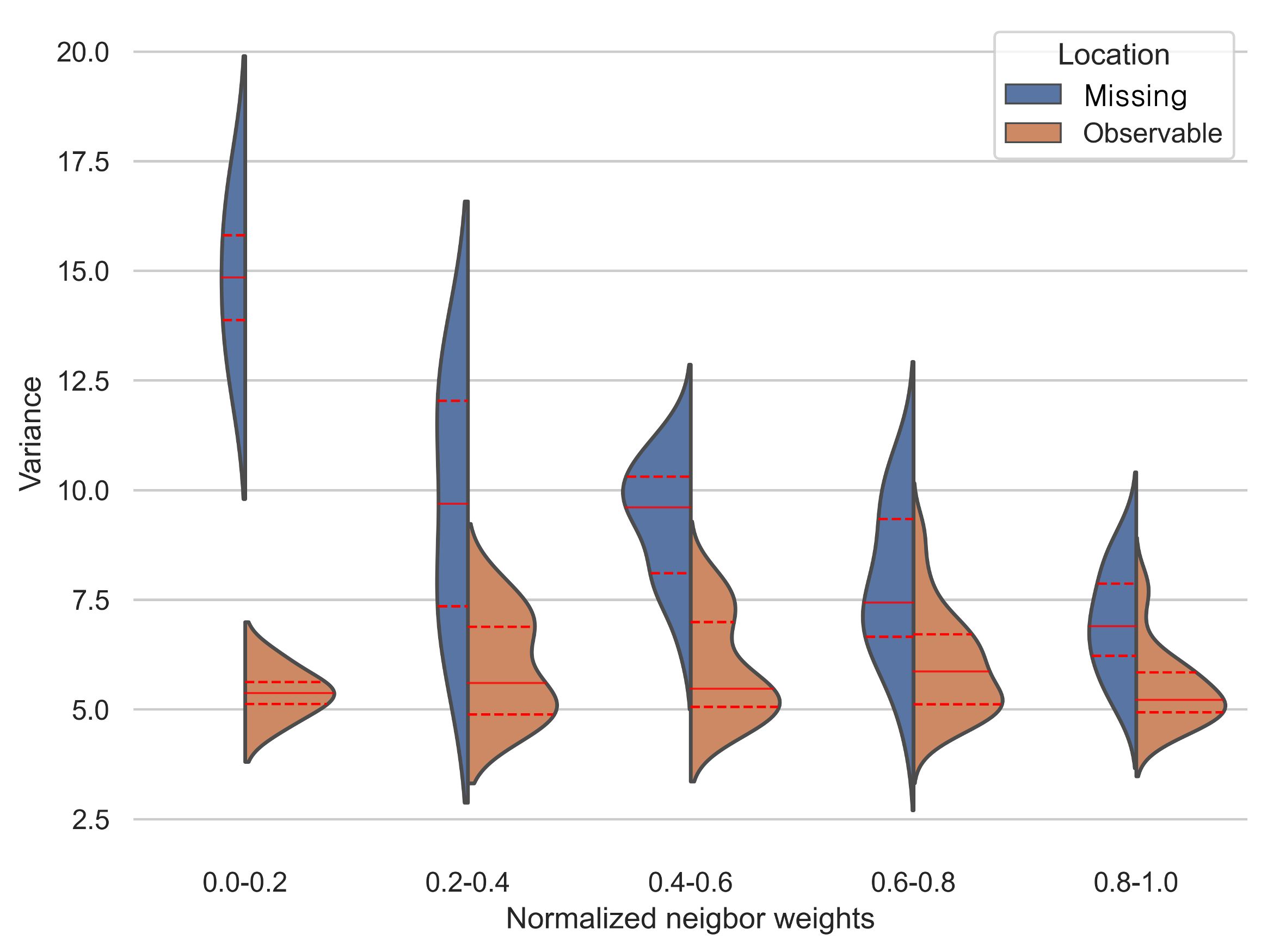}&
  \includegraphics[width=0.4\linewidth]{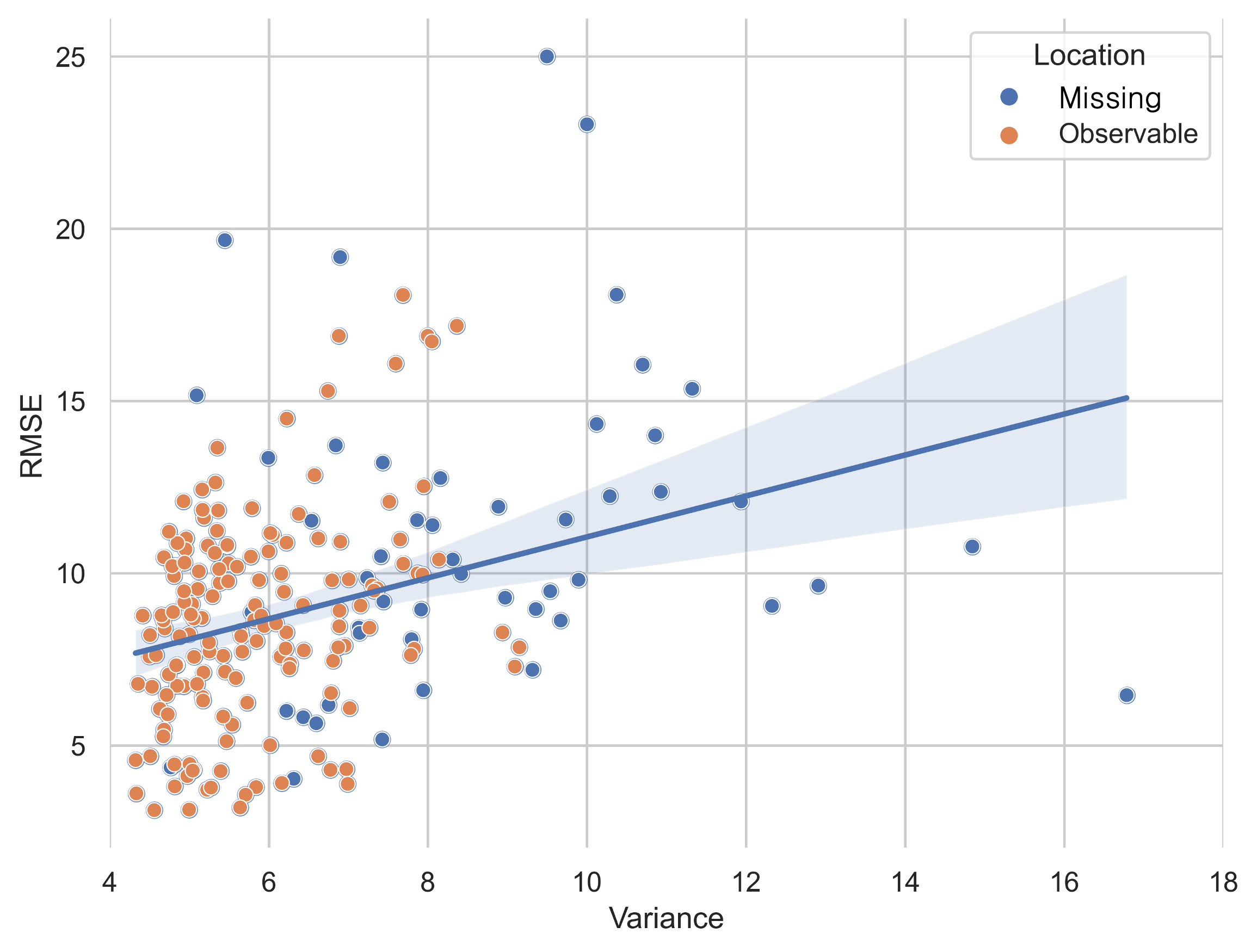}
  \\
  (c) Epistemic uncertainty distribution on different locations   & (d) Relationship between RMSE and epistemic uncertainty
  \end{tabular}
 \caption{Model uncertainty of 30 mins task prediction task on different locations of the METR-LA road network. (a) and (b) shows RMSE and NLL on different locations of the road network, and the starts and dots mean missing and observable locations. Compared to observable locations, missing locations have higher error and model uncertainty. (c) shows the Epistemic uncertainty of predictions at different locations. Missing locations with more and closer observable neighbors have lower model uncertainty, and observable locations do not have this effect. (d) shows that the error increases with the epistemic uncertainty increasing in both observable and missing locations.}
 \label{fig: RMSE NLL}
\end{figure*}

One of the distinct advantages of the proposed method is the capability to quantify uncertainties for its prediction.
To investigate the capability of uncertainty quantification of our method at different sensor locations, we conduct experiments on three traffic datasets and report NLL and RMSE regarding model uncertainty and accuracy at each location in Table~\ref{tab: performance of UIGNN}. The results show that observable ones have higher accuracy and less model uncertainty among all three datasets than missing locations. This is consistent with the detailed result on the METR-LA dataset shown in Fig.~\ref{fig: RMSE NLL} (a) and (b). Most of the missing locations (stars in the figure) have high RMSE and NLL compared to observable ones (dots in the figure). In Fig.~\ref{fig: RMSE NLL} (c), we use normalized neighbor weights which reflect the distance between the location and its surrounding observable neighbors. The larger the weight, the closer this location is to its nearby observable locations; the result shows the epistemic uncertainty at observable locations is not affected by its neighboring, while at missing locations, the epistemic uncertainty negatively correlates with its neighboring weights. This is due to the node feature at available observable locations, and our GNN model can learn better node representations. And at the missing locations, the result is weakened since these locations have no node features, and the node representations are aggregated from the neighboring locations. 

From Fig.~\ref{fig: RMSE NLL} (d), we found the uncertainty successfully reflects model performance at both observed and missing locations, as error steadily decreases with epistemic uncertainty decreasing.

\begin{table}[!ht]
\centering
\caption{Performance of UIGNN on 15 mins, 30 mins, and 1-hour prediction tasks on three real-world Datasets. RMSE and NLL reflect model accuracy and uncertainty, respectively.}
\label{tab: performance of UIGNN}
\centering
\setlength{\columnseprule}{20pt}
\resizebox{0.95\linewidth}{!}{
\begin{tabular}{c||ccccc}

\toprule\toprule

\multirow{2}{*} &
  \multirow{2}{*}{\textbf{Time}} &
  \multicolumn{2}{c}{\textbf{RMSE}} &
  \multicolumn{2}{c}{\textbf{NLL}}\\

 &
   &
  \multicolumn{1}{c}{\textbf{Observable}} &
  \textbf{Missing} &
  \multicolumn{1}{c}{\textbf{Observable}} &
  \textbf{Missing} \\
\midrule
  
\multirow{3}{*}{\rotatebox{90}{\textbf{METR-LA}}}  & \textbf{15 min} & \multicolumn{1}{c}{3.2874} & 7.1424 & \multicolumn{1}{c}{5.1535} & 8.6832 \\ \\[-1.5ex]
                                   & \textbf{30 min} & \multicolumn{1}{c}{3.9450} & 7.3052 & \multicolumn{1}{c}{5.5716} & 9.0121 \\  \\[-1.5ex]
                                   & \textbf{1 hour} & \multicolumn{1}{c}{5.0083} & 7.8019 & \multicolumn{1}{c}{6.3048} &  8.6404
                                   \\ 
                                   \\[-1.5ex]
\midrule

\multirow{3}{*}{\rotatebox{90}{\textbf{PeMS-BAY}}} & \textbf{15 min} & \multicolumn{1}{c}{1.6456} & 3.9945 & \multicolumn{1}{c}{3.5171} & 6.5895 \\ \\[-1.5ex]
                                   & \textbf{30 min} & \multicolumn{1}{c}{1.9735} & 4.1254 & \multicolumn{1}{c}{3.7642} & 7.0601 \\ 
                                   \\[-1.5ex]
                                   & \textbf{1 hour} & \multicolumn{1}{c}{2.7178} & 4.2870 & \multicolumn{1}{c}{5.1537} & 7.2614 \\ 
                                   \\[-1.5ex]
\midrule
\multirow{3}{*}{ \rotatebox{90}{\textbf{SeData}}}   & \textbf{15 min} & \multicolumn{1}{c}{3.1805} & 4.7916 & \multicolumn{1}{c}{6.2262} &  7.9262 \\ \\[-1.5ex]
                                   & \textbf{30 min} & \multicolumn{1}{c}{3.8114} & 4.9139 & \multicolumn{1}{c}{6.8129} & 8.1601 \\ \\[-1.5ex]
                                   & \textbf{1 hour} & \multicolumn{1}{c}{4.8856} & 5.8468 & \multicolumn{1}{c}{7.6020} & 8.5669 \\ \bottomrule

\end{tabular}}
\end{table}

\subsection{Case Study: Active Sensing Task (\textbf{RQ3})}

\begin{figure}[!t]
\centering
  \begin{tabular}{c}
  \includegraphics[width=0.8\linewidth]{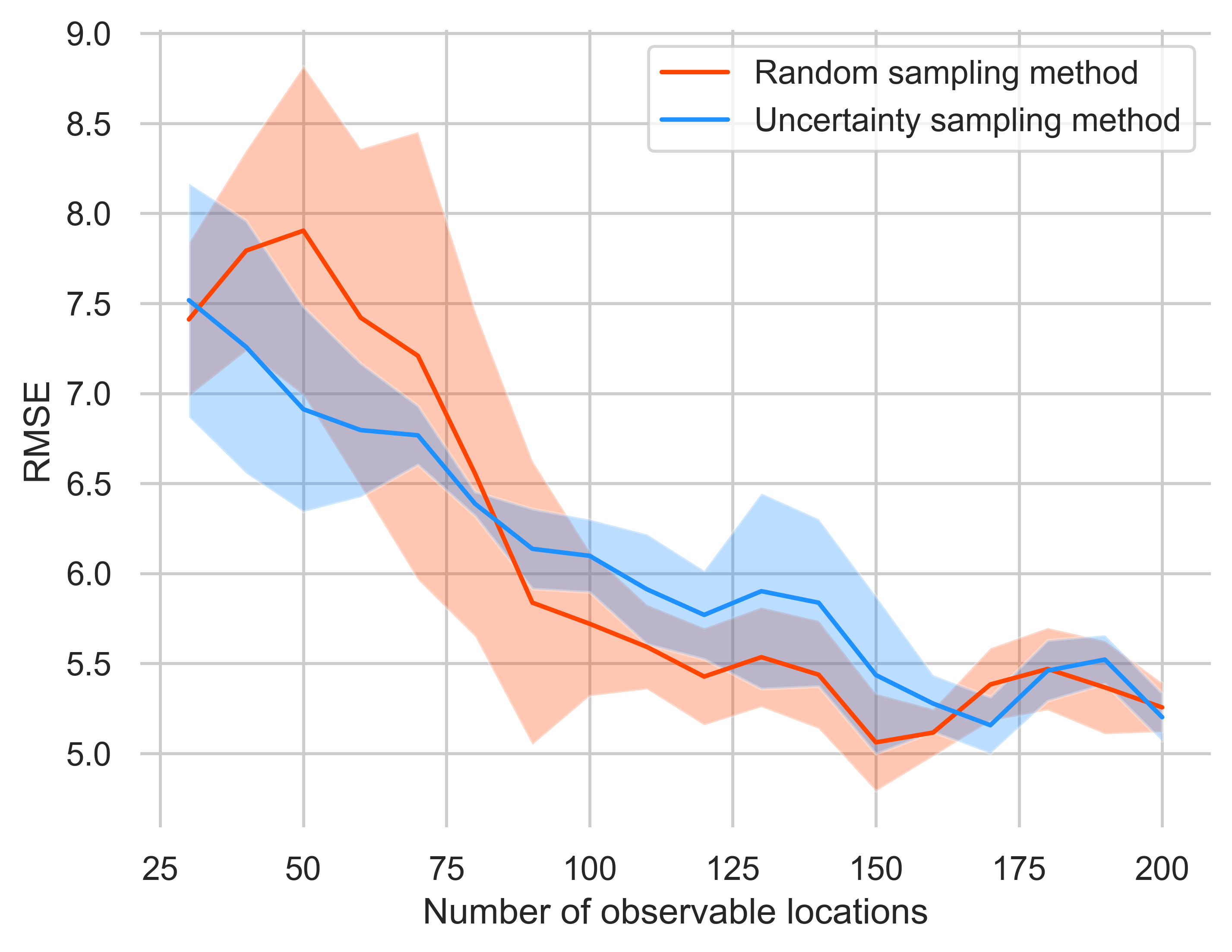}
  \\
  (a) RMSE of $\mathcal{V}_{o}$ at different $N_e$ on METR-LA
  \\
 \includegraphics[width=0.8\linewidth]{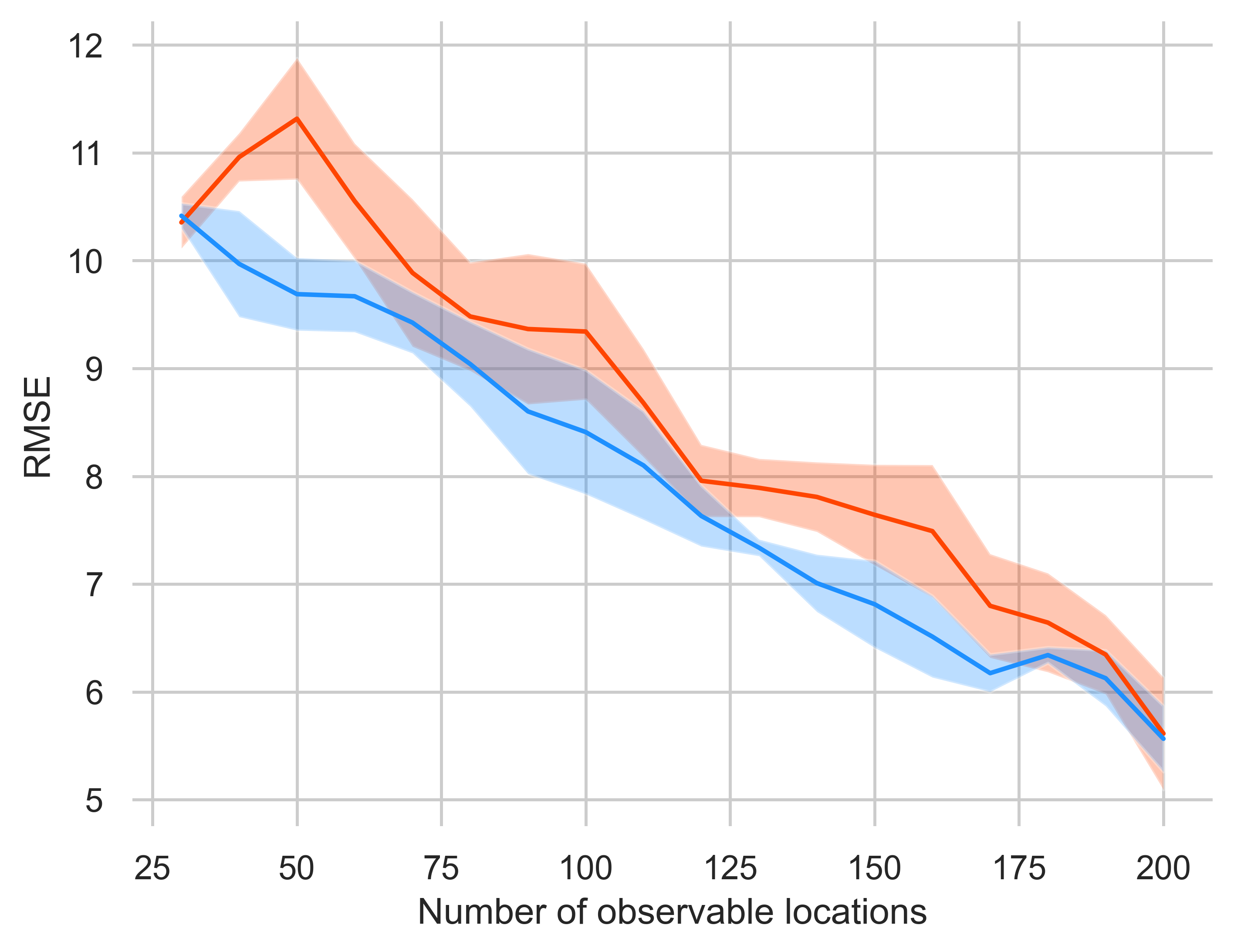}
  \\
  (b) RMSE of $\mathcal{V}_{e}$ at different $N_e$ METR-LA
  \end{tabular}
 \caption{Performance of Uncertainty sampling and random sampling method on active the sensing task. Lower RMSE means better prediction performance. (a) It shows different sampling methods have similar performance on observable locations. (b) The uncertainty sampling method performs better compared to the random sampling method. And two methods finally converge and achieve the same level of accuracy.}
 \label{fig: active}
\end{figure}

Fig.~\ref{fig: active} shows the performance of 1-hour traffic prediction on the METR-LA dataset under the uncertainty sampling method directed by the UIGNN model \textbf{(blue lines)} and random sampling method \textbf{(orange lanes)}:
\noindent\\ $\bullet$ In random and uncertainty sampling methods, the RMSE decreases with the number of observable locations increasing, and they finally achieve the same level of performance. This shows the performance of both methods benefits from more training data as the traffic sensing network becomes complete. And the same level of performance of different methods is also explainable as in the final few steps, the observable locations of the two methods are highly overlapping.
\noindent\\ $\bullet$ At unobserved locations, the RMSE of the uncertainty sampling method decreases faster than the random sampling method at the beginning. This is because, at high-uncertainty locations, the topology of the locations in the sensing network and traffic states of its surrounding locations are more likely to be out-of-distribution (OOD) samples which the learned model has difficulty generalizing to~\cite{hamilton2017inductive, amini2020deep}. Picking at these locations will make our model learn from the unseen data, thus decreasing the RMSE even if it is exposed to limited locations. And random sample method cannot benefit from uncertainty quantification at unobserved locations and is therefore exposed to these OOD data later than the uncertainty sampling method. As a result, the RMSE of this method decreases slower. 

\noindent\\ $\bullet$ At the observable locations, the RMSE of uncertainty and random sampling method have no major differences, and both of these methods achieve the best performance at around 100 observable locations. This is because of features at each location, and few of its neighbors can approximate the future traffic state accurately; thus, when nearly half of the locations are observable, the forecasting model achieves the best performance. Since uncertainty helps the sensing network pick up the most likely unseen data at high-uncertainty locations, it does not provide much improvement at locations with data already seen.


\section{Conclusion}

We investigate the traffic prediction problem in a realistic setting where data is missing at part of the locations. To extend prediction perception to these locations missing historical records, we propose an uncertainty-aware inductive graph neural network that can: 1) predict future traffic state at both observable locations and missing locations; 2) quantify the uncertainty of the prediction which can help evaluate the model performance at both observable and missing locations. We conduct extensive environments using three real-world datasets and testify our method performs better than intuitive solutions based on twp-steps (impute then forecast) approaches. In addition, we show in the case study on the METR-LA dataset that the uncertainty can help evaluate prediction accuracy and assist downstream tasks like sensor deployment on traffic networks.



\bibliographystyle{IEEEtran}
\bibliography{IEEEabrv,mybib}

\end{document}